%% file: main.tex
\documentclass{article}

\PassOptionsToPackage{numbers,sort&compress}{natbib}

\usepackage[preprint]{neurips_2025}

\usepackage[utf8]{inputenc} 
\usepackage[T1]{fontenc}    
\usepackage{hyperref}       
\usepackage{url}            
\usepackage{booktabs}       
\usepackage{amsfonts}       
\usepackage{nicefrac}       
\usepackage{microtype}      
\usepackage{xcolor}         

\title{Efficient Deployment of Vision-Language Models on Mobile Devices: A Case Study on OnePlus 13R}

\author{%
  Pablo Robin Guerrero%
    \thanks{Corresponding author: \texttt{pablo.robin@epfl.ch}}%
  \quad
  Yueyang Pan%
    \thanks{Supervisors}%
  \quad
  Sanidhya Kashyap\footnotemark[2]%
  \\[0.6ex]   
  École Polytechnique Fédérale de Lausanne (EPFL)
}

\input{macros}

\begin{document}

\maketitle
\input{tex/0_abstract}
\input{tex/1_intro}
\input{tex/2_related_work}
\input{tex/3_methodology}
\input{tex/4_results}
\input{tex/5_discussion}
\input{tex/6_conclusion}

\clearpage

\begin{ack}
\input{tex/7_acknowledgement}
\end{ack}

\bibliography{main}
\bibliographystyle{unsrtnat}

\end{document}

%% file: macros.tex
\usepackage{xspace}
\usepackage{graphicx}
\usepackage{amsmath}
\usepackage{amssymb}
\usepackage{amsthm}
\usepackage{adjustbox}
\usepackage{wrapfig}
\usepackage{lipsum}
\usepackage{booktabs}
\usepackage{multirow}
\usepackage{makecell}

\usepackage{algorithm}
\usepackage{algpseudocode} 

\algrenewcommand\algorithmicrequire{\textbf{Input:}}
\algrenewcommand\algorithmicensure{\textbf{Output:}}

\usepackage{subcaption}     
\usepackage{subcaption}

\newcommand{\xhdr}[1]{{\noindent\bfseries #1}.}

\newcommand{\rom}[1]{\textup{\uppercase\expandafter{\romannumeral#1}}}

\newcounter{observation}

\newcounter{design}

\usepackage[most]{tcolorbox}
\usepackage{colortbl}
\usepackage{pifont}
\usepackage{makecell}

\tcbuselibrary{skins, breakable}   
\usepackage{tabularx, booktabs}    

\definecolor{darkred}{rgb}{0.5,0,0}
\definecolor{darkgreen}{rgb}{0,0.5,0}

\newtcolorbox[
  auto counter,
  list inside=examplelist,
]{greenbox}[2][]{
  enhanced,
  title={Example~\thetcbcounter. \textbf{#1}},
  colback=green!5!white,
  colframe=green!50!black,
  colbacktitle=green!60!black,
  coltitle=white,
  #2
}

\newtcolorbox[
  auto counter,
  list inside=examplelist
]{bluebox}[2][]{
  enhanced,
  title={Text~\thetcbcounter. \textbf{#1}},
  colback=cyan!5!white,  
  colframe=cyan!50!black,  
  colbacktitle=cyan!75!black,  
  coltitle=white,
  #2
}

\usepackage{pifont}


%% file: tex/0_abstract.tex
\begin{abstract}
\label{sec:abstract}
Vision-Language Models (VLMs) offer promising capabilities for mobile devices, but their deployment faces significant challenges due to computational limitations and energy inefficiency, especially for real-time applications.
This study provides a comprehensive survey of deployment frameworks for VLMs on mobile devices, evaluating \texttt{llama.cpp}, MLC-Imp, and \texttt{mllm} in the context of running LLaVA-1.5 7B, MobileVLM-3B, and Imp-v1.5 3B as representative workloads on a OnePlus 13R.
Each deployment framework was evaluated on the OnePlus 13R while running VLMs, with measurements covering CPU, GPU, and NPU utilization, temperature, inference time, power consumption, and user experience.
Benchmarking revealed critical performance bottlenecks across frameworks: CPU resources were consistently over-utilized during token generation, while GPU and NPU accelerators were largely unused. When the GPU was used, primarily for image feature extraction, it was saturated, leading to degraded device responsiveness.
The study contributes framework-level benchmarks, practical profiling tools, and an in-depth analysis of hardware utilization bottlenecks, highlighting the consistent overuse of CPUs and the ineffective or unstable use of GPUs and NPUs in current deployment frameworks.
\end{abstract}

%% file: tex/1_intro.tex
\section{Introduction}
Vision-Language Models (VLMs) have achieved significant progress in solving multimodal tasks such as visual question answering, image captioning, and image-grounded reasoning by combining a visual encoder with a large language model to process both image and text inputs. Recent advances have enabled these models to produce increasingly accurate and coherent responses across complex visual-language benchmarks \citep{densefusion2024, lma2024survey}. However, such improvements come at the cost of substantial computational overhead, making VLMs difficult to deploy on resource-constrained platforms such as mobile devices \citep{powerinfer2024, empowering2024}.

Despite these challenges, deploying VLMs on mobile devices offers four key advantages: enhanced privacy, support for diverse real-time applications, low-latency interaction, and fully offline operation. These capabilities are particularly valuable in contexts involving sensitive data, unreliable connectivity, or latency-critical tasks. Additionally, on-device inference enables energy-aware execution, making it well-suited for power-constrained environments. However, delivering these benefits requires overcoming the severe hardware limitations of mobile platforms, including restricted compute capacity, limited memory bandwidth, and tight thermal budgets \citep{swapmoe2023, empowering2024}.

Efficiently deploying VLMs on mobile hardware presents non-trivial challenges due to the need to map distinct stages of the inference pipeline, such as image encoding, feature fusion, and token generation, onto heterogeneous compute units including CPUs, GPUs, and NPUs. Image encoding and attention layers are typically GEMM-dominated and benefit from GPU or NPU acceleration. In contrast, token generation involves sequential decoding with GEMV-heavy computation, which exhibits low parallelism and is often offloaded to CPUs. NPUs are particularly inefficient for GEMV workloads due to limited data reuse and low arithmetic intensity \citep{empowering2024, powerinfer2024}. These heterogeneous workload characteristics complicate scheduling and must be managed within tight thermal, memory, and power constraints. As a result, delivering responsive and energy-efficient inference on mobile devices remains a fundamentally difficult systems problem.

To address these challenges, a wide range of techniques has been developed for edge optimization, including quantization, pruning, distillation, and memory-efficient attention mechanisms. These methods aim to reduce model size, latency, and energy consumption while preserving task accuracy. In parallel, several deployment frameworks have emerged to enable vision-language inference on mobile devices. Among them, \texttt{llama.cpp}~\citep{llamacpp2023}, MLC-Imp~\citep{mlcimp2024}, and \texttt{mllm}~\citep{mllm2024} adopt different strategies for leveraging mobile hardware capabilities, supporting various levels of model optimization and accelerator integration. However, it remains unclear how these frameworks perform under real-world constraints such as compute heterogeneity, memory limits, and power budgets. The lack of systematic empirical evaluation makes it difficult to assess trade-offs and guide practitioners in selecting the most suitable framework for specific models and deployment scenarios.

We conduct the first cross-framework evaluation of VLM deployment on mobile devices, analyzing how four representative runtimes (namely \texttt{llama.cpp}~\citep{llamacpp2023}, MLC-Imp~\citep{mlcimp2024}, and \texttt{mllm}~\citep{mllm2024}) perform under real hardware constraints. Using LLaVA-1.5 7B~\citep{llava2024}, MobileVLM-3B~\citep{mobilevlm2024}, and Imp-v1.5 3B~\citep{imp2024} as representative workloads, we benchmark each framework on a modern smartphone (OnePlus 13R) across a range of system-level metrics, including CPU, GPU, and NPU utilization, inference latency, power consumption, thermal behavior, and user experience.

Our evaluation reveals consistent overutilization of CPU resources during token generation, along with underutilization or ineffective use of GPU and NPU accelerators. In several cases, attempts to offload image processing to the GPU led to device instability, including screen freezes and degraded responsiveness~\citep{empowering2024}.
To support reproducibility and future research, we release an open-source benchmarking pipeline and detailed profiling results. Our findings expose key limitations in current mobile VLM deployment pipelines and motivate the development of improved hardware-aware scheduling and accelerator utilization strategies~\citep{powerinfer2024, swapmoe2023}.

%% file: tex/2_related_work.tex

\section{Related Work}

\xhdr{Model Optimization for Edge Deployment}
Various techniques have been employed to reduce the resource demands of large models on constrained hardware. Quantization~\citep{frantar2022gptq, lin2023awq}, pruning~\citep{han2015pruning}, distillation~\citep{hinton2015distillation}, and operator-level optimizations such as FlashAttention~\citep{dao2022flashattention} are widely adopted in mobile inference pipelines. However, their collective impact remains unclear, as existing deployment frameworks integrate them to varying degrees without consistent evaluation. In this work, we treat these optimizations as part of the runtime environment and focus on end-to-end deployment behavior under real-world mobile constraints.

\xhdr{Vision-Language Benchmarks and Models}
Recent VLMs such as LLaVA~\citep{llava2024}, MobileVLM~\citep{mobilevlm2024}, and DeepSeek-VL 2~\citep{deepseekvl2_2024} have demonstrated strong performance on multimodal tasks. While MobileVLM is specifically designed for mobile deployment, most vision-language models are evaluated primarily on GPU clusters, with limited attention to efficiency on resource-constrained hardware. Existing studies emphasize accuracy, often overlooking system-level metrics such as latency, power consumption, or hardware utilization. Our work addresses this gap by evaluating VLMs from a deployment perspective on consumer devices.

\xhdr{Edge and Mobile Inference Frameworks}
Multiple runtimes now support on-device inference, including \texttt{llama.cpp}~\citep{llamacpp2023}, MLC-Imp~\citep{mlcimp2024}, and \texttt{mllm}~\citep{mllm2024}, each with varying levels of compatibility for quantization, accelerator offloading, and runtime flexibility. Prior studies such as PowerInfer~\citep{powerinfer2024} and mllm-NPU~\citep{empowering2024} explore individual frameworks, but no comparative evaluation exists for vision-language workloads. This study fills that gap by benchmarking frameworks and models jointly, capturing deployment-relevant system metrics on real mobile hardware.

%% file: tex/3_methodology.tex
\section{Methodology}

\subsection{Evaluation Goals}
The goal of this study is to evaluate the deployment efficiency of vision-language models (VLMs) on a OnePlus 13R smartphone. Specifically, we examine how existing deployment frameworks perform when running representative multimodal workloads under real-world constraints. Our evaluation targets deployment-relevant metrics, including inference latency, CPU/GPU/NPU utilization, power consumption, thermal behavior, and user responsiveness. This analysis aims to identify system bottlenecks, compare framework-level trade-offs, and inform future work on efficient on-device multimodal inference.

\subsection{Models and Frameworks}
We evaluate three representative vision–language models: LLaVA-1.5 7B~\citep{llava2024}, MobileVLM-3B~\citep{mobilevlm2024}, and Imp-v1.5 3B~\citep{imp2024}. LLaVA-1.5-7B provides a high-capacity baseline, MobileVLM-3B targets fully on-device inference, and Imp-v1.5-3B distils LLaVA-1.5 into a 3 B-parameter model built on a Phi-2 core with a SigLIP visual encoder.

These models are deployed using four mobile-friendly inference frameworks: \texttt{llama.cpp}~\citep{llamacpp2023}, MLC-Imp~\citep{mlcimp2024}, and \texttt{mllm}~\citep{mllm2024}. We evaluate each model with all compatible frameworks to capture cross-runtime performance differences under realistic deployment conditions.

\subsection{Hardware and Runtime Environment}
All experiments were conducted on a OnePlus 13R smartphone running Android 15. The device is powered by a Qualcomm Snapdragon 8 Gen 2 chipset, which includes an octa-core Kryo CPU, Adreno 740 GPU, and a Hexagon NPU. The test model features 12 GB of LPDDR5X RAM and 256 GB of UFS 4.0 storage.

To enable fine-grained system monitoring, the device was rooted to access low-level performance counters and logging interfaces. Frameworks were executed in native Android environments using official or community-supported builds. Measurements were collected using a combination of system profiling tools, custom logging scripts, and screen recordings to ensure consistency across runs.

\subsection{Evaluation Metrics}
To assess deployment efficiency, we report a range of system- and user-level metrics relevant to real-time mobile inference, as summarized in Table~\ref{tab:eval_metrics}.

\begin{table}[t]
  \caption{Evaluation metrics for on-device VLM deployment}
  \vspace{2pt}
  \label{tab:eval_metrics}
  \centering
  \begin{tabularx}{\linewidth}{l|X}
    \toprule
    \textbf{Metric} & \textbf{Description} \\
    \midrule
    Inference latency & End-to-end time from input to output, measured separately for image encoding and token generation when applicable. \\[3ex]
    Hardware utilization & Real-time CPU, GPU, and NPU usage recorded throughout inference to assess workload distribution and detect resource bottlenecks. \\[3ex]
    Power consumption & Instantaneous and average power draw measured during inference, used to estimate energy impact and battery drain. \\[3ex]
    Thermal behavior & Surface temperature tracked over time to evaluate thermal buildup, potential throttling, and thermal stability. \\[3ex]
    User responsiveness & Qualitative assessment of device interactivity during inference. Includes UI lag or screen freezing for graphical interfaces, and I/O delay or shell blocking for command-line usage. \\
    \bottomrule
  \end{tabularx}
\end{table}

\subsection{Benchmarking Procedure}
For each model–framework combination, we evaluate inference performance using standardized prompts and input images. To reduce variability, we disable automatic screen brightness and run all experiments in airplane mode with background processes minimized. Each benchmark is repeated five times, and we report average values across runs.

Inference is segmented into four phases: \textit{load} (model or input preparation), \textit{slice-encode} (image preprocessing), \textit{prompt-eval} (context embedding), and \textit{token-eval} (autoregressive decoding). These phases are annotated in our //usage plots to highlight temporal workload distribution.

System activity is recorded using custom monitoring scripts that log CPU, GPU, and NPU utilization at 100ms intervals via root-accessible system files under \texttt{/proc} and \texttt{/sys}. Power consumption and thermal behavior are captured from vendor-specific sensors exposed through \verb|/sys/class/power_supply/| and thermal zones. Logging runs in parallel with inference, and timestamps are aligned with screen recordings to synchronize system measurements with user-perceived events.

%% file: tex/4_results.tex
\section{Results}

\subsection{Overview}

Our evaluation reveals substantial performance and system behavior variations across the tested deployment frameworks. While all frameworks successfully executed the selected models on-device, their efficiency varied significantly in latency, hardware utilization, and energy impact.

Inference consistently bottlenecked at the CPU during token generation, with the GPU mostly underutilized and the Hexagon NPU entirely unused. This underutilization notably impacts battery life and device responsiveness. Frameworks offering partial accelerator support (e.g., GPU-only image encoding) occasionally caused device instability, such as screen freezing during GPU load spikes.

\subsection{Latency Analysis}

We analyze latency by decomposing the vision-language pipeline into four stages: model loading, image encoding, prompt evaluation, and autoregressive token generation.

For \textbf{LLaVA-1.5 7B}, Table~\ref{tab:latency_combined} illustrates that \texttt{llama.cpp} required approximately 3.1 seconds for image encoding, while prompt evaluation dominated runtime, taking 89 seconds for 605 tokens (147.2 ms/token). Token generation added 11.8 seconds for 69 tokens (171.7 ms/token), resulting in over 101 seconds total latency.

Conversely, \texttt{mllm} completed image encoding rapidly (38 ms) but encountered severe delays during prompt evaluation (78 seconds for 19 tokens, ~4,153 ms/token). Token generation was similarly slow (94 seconds for 51 tokens, ~1,860 ms/token). The discrepancy between \texttt{mllm}'s reported internal throughput (4,858 tokens/sec) and observed latency underscores significant inefficiencies in scheduling and accelerator utilization.

These results emphasize two key observations: (1) prompt evaluation is the dominant latency bottleneck for LLaVA-1.5 7B, and (2) framework-level implementation choices significantly influence runtime efficiency, even for the same underlying model.

For \textbf{Mobile-scale models}, \texttt{llama.cpp} running MobileVLM-3B showed faster overall visual preprocessing (~14.1 seconds) compared to \texttt{MLC-Imp} with Imp-v1.5-3B (~18 seconds). Despite faster prompt evaluation (~2 s) and decoding (~1 s), the longer preprocessing and model load in \texttt{MLC-Imp} increased total latency (~25 s) compared to \texttt{llama.cpp} (~21 s). This highlights runtime-level optimization opportunities, as Imp-v1.5-3B does not outperform MobileVLM-3B on visual latency, and both remain CPU-bound during token generation.

\begin{table}[t]
  \caption{Latency breakdown (milliseconds) for LLaVA-1.5 7B, MobileVLM-3B, and \textsc{Imp-v1.5-3B} across frameworks. ‘—’ means the framework does not expose that timing.  Totals for \texttt{llama.cpp} rows are the wall-clock “total time” printed by the binary.}
  \label{tab:latency_combined}
  \centering
  \begin{tabularx}{\linewidth}{
    l|
    >{\centering\arraybackslash}X
    >{\centering\arraybackslash}X|
    >{\centering\arraybackslash}X
    >{\centering\arraybackslash}X
  }
    \toprule
    \multicolumn{1}{c|}{} &
    \multicolumn{2}{c|}{\textbf{LLaVA-1.5 7B}} &
    \multicolumn{2}{c}{\textbf{MobileVLM-3B / \textsc{Imp-v1.5-3B}}} \\
    \cmidrule(lr){2-3}\cmidrule(lr){4-5}
    \textbf{Stage} & \texttt{llama.cpp} & \texttt{mllm} &
                    \texttt{llama.cpp} & \texttt{MLC-Imp} \\
    \midrule
    Model load & 3209.1 & 4479.7 & 2334.4 & 4000.0 \\
    Image encoding & 2399.0 & 38.5 & 3053.0 & 18000.0 \\
    Image decoding & 63724.0 & — & 9989.0 & — \\
    Prompt evaluation & 70391.4 & 78901.5 & 15300.2 & 2000.0 \\
    Token generation & 11085.8 & 90249.0 & 6605.2 & 1000.0 \\
    \midrule
    \textbf{Total} &
    \textbf{82194.9} & \textbf{173668.7} &
    \textbf{22799.1} & \textbf{25000.0} \\
    \bottomrule
  \end{tabularx}
\end{table}

\subsection{Hardware Utilization}
\label{subsec:hw-util}

\begin{figure}[t]
  \centering

  \begin{subfigure}[t]{\columnwidth}
    \includegraphics[width=0.9\linewidth]{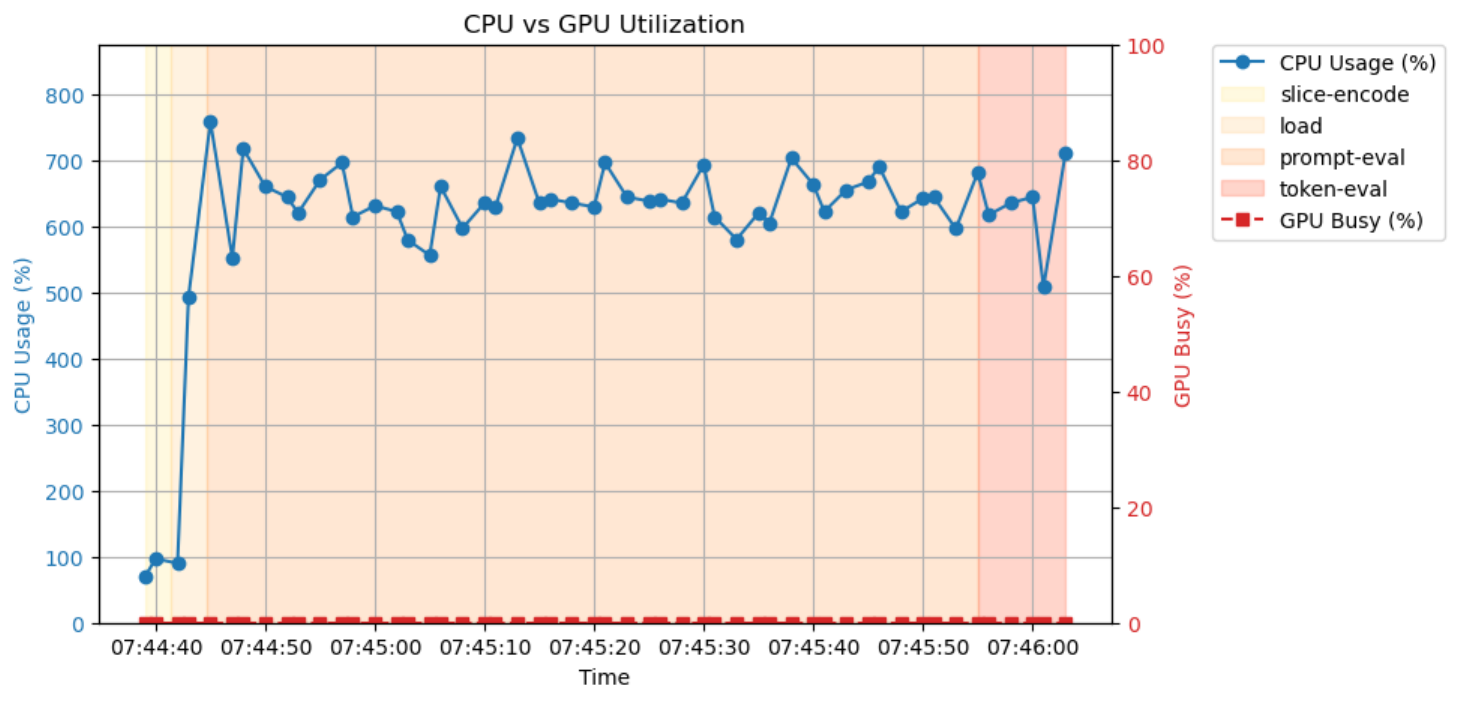}
    \caption{\texttt{llama.cpp}}
    \label{fig:hwutil-llava-llamacpp}
  \end{subfigure}

  \vspace{0.8\baselineskip}

  \begin{subfigure}[t]{\columnwidth}
    \includegraphics[width=0.9\linewidth]{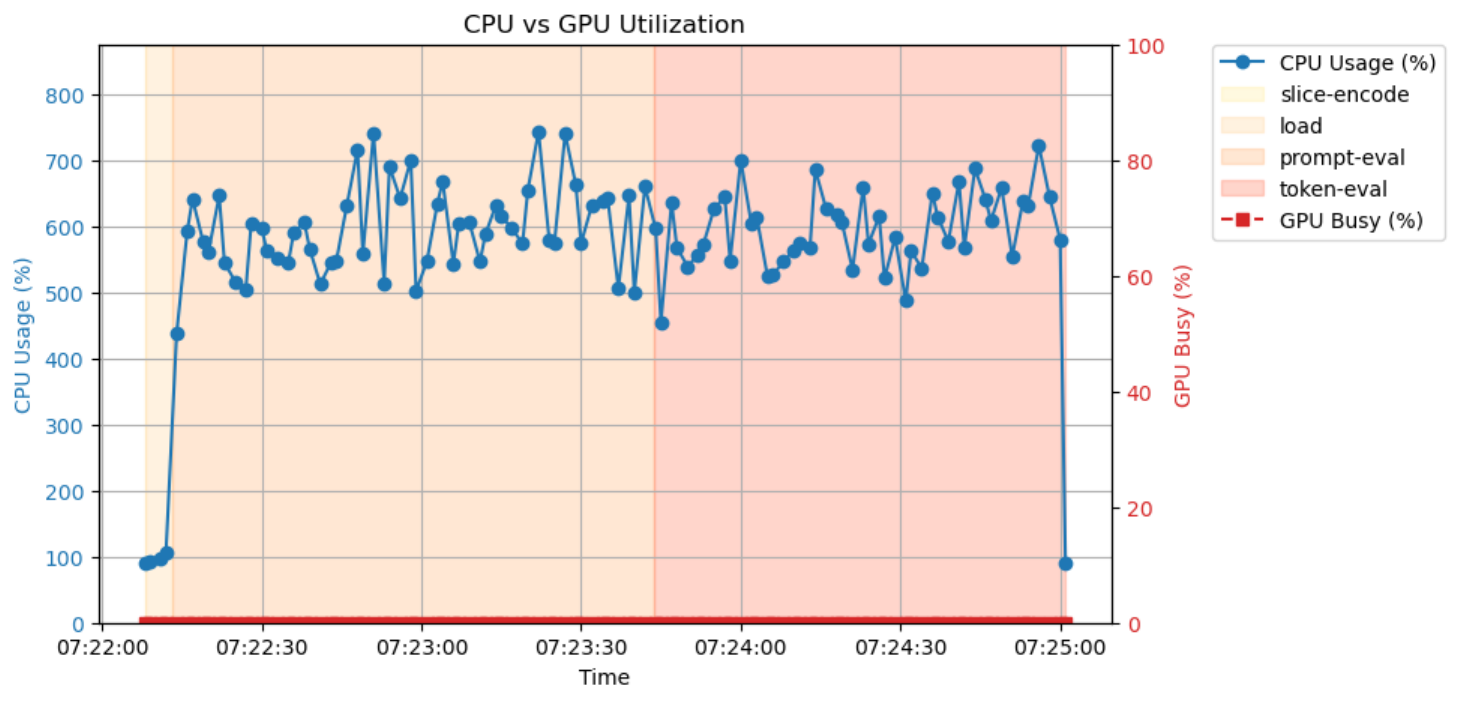}
    \caption{\texttt{mllm}}
    \label{fig:hwutil-llava-mllm}
  \end{subfigure}

  \caption[CPU \& GPU utilisation for LLaVA-1.5 7B]{%
    CPU--GPU utilisation while running \textsc{LLaVA-1.5 7B} on a One\-Plus~13R.%
    \textbf{Left:} \texttt{llama.cpp} drives CPU load to $\approx 600\,\%$
    (8~cores) for most of the $90\,\text{s}$ run; the Adreno~740 stays idle.%
    \textbf{Right:} \texttt{mllm} peaks even higher ($\approx 750\,\%$) yet still
    leaves the GPU unused.%
    Shaded regions mark pipeline stages (\emph{slice-encode}, \emph{load},
    \emph{prompt-eval}, \emph{token-eval}).}
  \label{fig:hwutil-llava-side}
\end{figure}

During the \textsc{LLaVA-1.5 7B} deployments, both \texttt{llama.cpp} and \texttt{mllm} demonstrated fully CPU-bound workloads without GPU/NPU use, causing significant thermal stress (Figs.~\ref{fig:hwutil-llava-llamacpp}–\ref{fig:hwutil-llava-mllm}). \texttt{llama.cpp} a \(\sim\!200\text{ ms}\) image-encoding burst is followed by prompt evaluation (\(\sim\!80\text{ s}\)) and decoding (\(\sim\!10\text{ s}\)); throughout these stages  maintained high CPU utilization (~600\%) of the eight cores, while the Adreno 740 GPU stays at \(\mathbf{0\,\%}\) and the Hexagon NPU is never invoked. Die temperature climbs from \(\sim\!40^{\circ}\text{C}\) to a \(\mathbf{88\text{–}90^{\circ}\text{C}}\) plateau within \(\mathbf{15}\text{ s}\). Battery current rises from \(\mathbf{0.2}\text{ A}\) at idle to \(\sim\!\mathbf{3.1}\text{ A}\) (\(\approx\!\mathbf{10}\text{ W}\)), and memory use settles at \(\sim\!\mathbf{55\,\%}\) of the 16 GB pool.

Under \texttt{mllm} the pattern intensifies: after a similar model-load phase, CPU utilisation oscillates between \textbf{650\,\% and 750\,\%} for the entire \(\sim\!180\)\,s run, while the GPU again remains at \textbf{0\,\%} busy and the NPU idle. Sustained power draw holds at
\(\approx\)\,\textbf{2.9 A} (\(\approx\)\,\textbf{11 W}), pushing die temperature slightly higher, to a \textbf{92–95\,\(^{\circ}\mathrm{C}\)} plateau. Prompt evaluation is roughly \textbf{2×} slower than in \texttt{llama.cpp}, and memory tops out at \(\sim\!\)\textbf{40\,\%}, reflecting a larger in-memory KV cache.

Taken together, the results show that neither framework exploits the mobile GPU or NPU; instead, both push the CPU to its thermal envelope, with \texttt{mllm} trading still higher temperature and power for no latency benefit.

\begin{figure}[t]
  \centering
  \includegraphics[width=0.9\linewidth ]{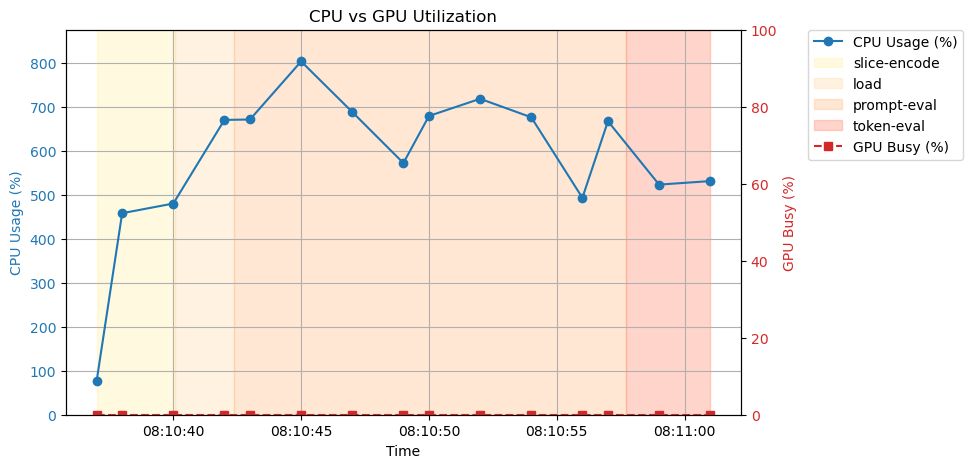}
  \caption{CPU versus GPU utilisation while running \textsc{MobileVLM-3B} under \texttt{llama.cpp} on the OnePlus 13R. During prompt evaluation and decoding, CPU load ranges from 500–800 \% of eight cores, whereas the Adreno 740 GPU remains idle (\(0\,\%\)). Shaded bands denote the four pipeline stages (\textit{slice-encode}, \textit{image-decode}, \textit{prompt-eval}, \textit{token-gen}).}
  \label{fig:hwutil-mobilevlm-llamacpp}
\end{figure}

The MobileVLM-3B deployment on \texttt{llama.cpp} also showed a fully CPU-bound workload but lower thermal impact (70–72°C) and reduced memory usage (~32\%) (Fig.~\ref{fig:hwutil-mobilevlm-llamacpp}). After slice-encode (3 s) and image-decode (10 s), CPU load stays at \(\mathbf{500\text{–}800\,\%}\) of eight cores for 15 s of prompt evaluation and 7 s of decoding, while the Adreno 740 remains at \(\mathbf{0\,\%}\) busy.  Die temperature levels off at \(\mathbf{70\text{–}72^{\circ}\mathrm{C}}\), about 18 °C cooler than the LLaVA trace, and \texttt{dumpsys} reports a flat \(\sim\!57\text{ mA}\) rail current (orders of magnitude below the >2 A spikes seen with LLaVA).  Memory plateaus at \(\sim\!\mathbf{32\,\%}\) of 16 GB (\(\approx\!5.1\text{ GB}\)).  End-to-end latency is \(\sim\!35\text{ s}\); pushing the ViT encoder and attention blocks onto the Adreno GPU or Hexagon NPU could cut that time and power draw substantially.

\begin{figure}[t]
  \centering
  \includegraphics[width=0.9\linewidth]{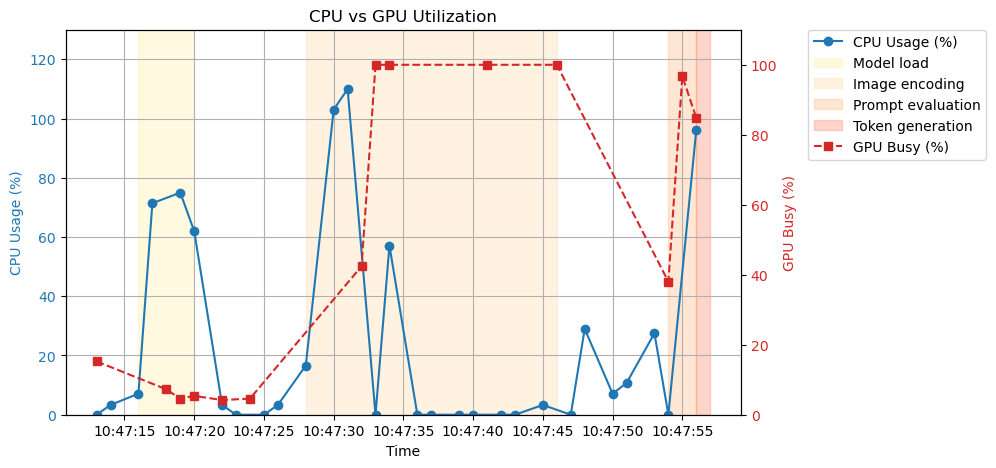}
  \caption{CPU versus GPU utilisation while running \textsc{Imp-v1.5-3B} under \texttt{MLC-Imp} on the OnePlus 13R. During the 18 s image-encoding phase and the subsequent 3 s text stack, the Adreno 740 is saturated above 90 \%, whereas aggregate CPU usage peaks at only ~120 \% of eight cores. Shaded bands mark pipeline stages (\textit{model load}, \textit{image encoding}, \textit{prompt evaluation}, \textit{token generation}).}
  \label{fig:hwutil-imp-mlc}
\end{figure}

The Imp-v1.5-3B model running on \texttt{MLC-Imp} successfully utilized the GPU, showing low CPU load (~120\%) (Fig.~\ref{fig:hwutil-imp-mlc}), significantly lower temperatures (~60°C), and reduced power draw (~1.3 W). Despite these advantages, overall latency was still limited by sequential token generation on the CPU, highlighting opportunities for further optimization. Unshaded intervals in the trace correspond to user interaction, photo selection and prompt typing, and are excluded from stage timings.  After a brief 4 s model-load phase, image encoding lasts \(\sim\!18\text{ s}\); over this window the Adreno 740 climbs from 40 \% to a sustained \textbf{100 \% busy} while CPU utilisation never exceeds 80 \%.  In the subsequent \(\sim\!2\text{ s}\) prompt-evaluation and 1 s decoding windows, the GPU remains above 90 \%, whereas aggregate CPU load tops out at just \textbf{120 \% of eight cores} (\(\approx\) one big core plus background threads). The lighter CPU load holds die temperature to a modest \textbf{60 \(^{\circ}\mathrm{C}\) peak}, about 30 \(^{\circ}\mathrm{C}\) cooler than the \texttt{llama.cpp}+LLaVA run, and battery current hovers near \(\sim\!0.33\text{ A}\) (\(\approx\!1.3\text{ W}\)), with only minor spikes when GPU utilisation surges.  Memory pressure is negligible, rising to \(<3\,\%\) of the 16 GB pool and flat-lining thereafter.  These traces confirm that \texttt{MLC-Imp} successfully off-loads both vision and language kernels to the GPU, alleviating the CPU bottleneck and delivering cooler, lower-power operation, yet token generation remains sequential, so the overall latency is still bounded by one-token-at-a-time decoding.

Taken together, Taken the phone remains \emph{entirely CPU-bound}: the GPU/NPU sit idle, CPU load peaks at \(800\,\%\) of eight cores, die temperature settles near \(71^{\circ}\text{C}\), and the query completes in \(\sim\!35\text{ s}\). In contrast, \texttt{MLC-Imp}+Imp-v1.5-3B off-loads nearly all vision–language compute to the Adreno 740, holding the CPU to \(\le120\,\%\) load, capping temperature at \(60^{\circ}\text{C}\), and drawing only \(0.33\,\text{A}\), yet overall latency is still bounded by one-token-at-a-time decoding. The comparison highlights two main levers for future speed-ups: (i) off-loading encoders and attention blocks to the GPU or NPU, and (ii) accelerating autoregressive decoding itself.

\subsection{Energy and Thermal Impact}

\begin{figure}[t]
  \centering
  \includegraphics[width=0.7\linewidth]{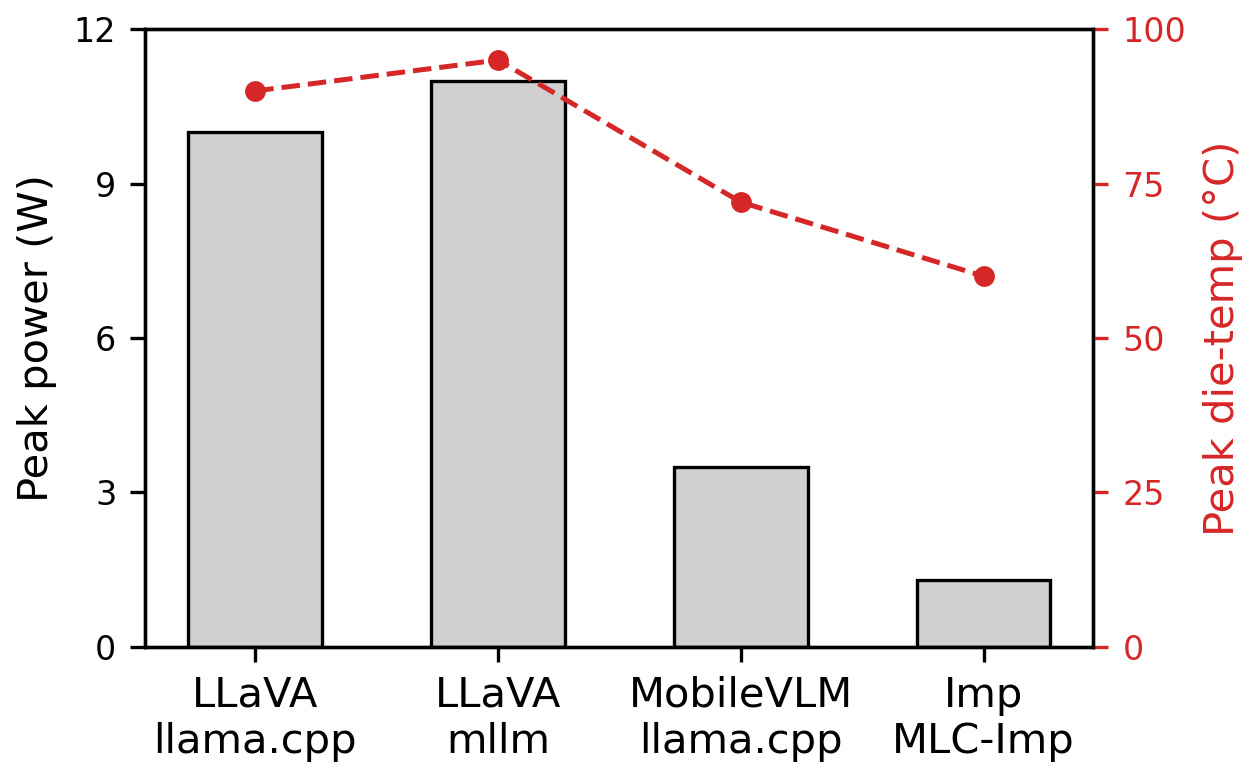}
  \caption{Peak package power (left axis) and die temperature (right axis) for each framework–model pair. CPU-only LLaVA stacks consume an order of magnitude more power and run 30 °C hotter than the GPU-off-loaded \texttt{MLC-Imp} path.}
  \label{fig:power-temp-summary}
\end{figure}

Significant differences emerged in power consumption and thermal profiles (Fig.~\ref{fig:power-temp-summary}). CPU-only LLaVA stacks were notably hotter and more power-intensive (10–12 W, 90–95°C). GPU-offloaded deployments showed substantial improvements, with MobileVLM-3B reaching ~3.5 W and 72°C, and Imp-v1.5-3B achieving even lower values at 1.3 W and 60°C. Practical implications of these differences include longer battery life, improved device longevity, and reduced user discomfort from thermal throttling. At a workload of one query per minute these figures translate to roughly eight hours of battery life for CPU-only LLaVA versus almost two days for the GPU path, underscoring the need to map encoder and attention kernels to the Adreno GPU or Hexagon NPU for sustained mobile use.

\subsection{Qualitative Output Comparison}

Model size directly correlated with output brevity. Both LLaVA stacks (\texttt{llama.cpp} and \texttt{mllm}) provided identical detailed three-sentence captions identifying “bacon, lettuce, tomatoes” on a bun. In contrast, \textbf{MobileVLM-3B on \texttt{llama.cpp}} condensed this to two sentences (“close-up of a burger with lettuce, tomatoes, and cheese”), while the smaller \textbf{Imp-v1.5-3B on \texttt{MLC-Imp}} offered a concise single-clause description (“a large hamburger with melted cheese and a toasted bun”). This varying brevity can significantly benefit resource-constrained applications requiring rapid response times or minimal communication overhead, with all models accurately classifying the object\footnote{See Table~\ref{tab:answer-length} for exact token counts.}.

\begin{table}[t]
  \captionsetup{skip=6pt}
  \centering
  \caption{Average answer length generated by each model.}
  \label{tab:answer-length}
  \begin{tabular}{@{}l r@{}}
    \toprule
    Model & Answer length (tokens) \\
    \midrule
    LLaVA+llama.cpp & 30 \\
    LLaVA+mllm & 30 \\
    MobileVLM & 18 \\
    Imp (MLC-Imp) & 11 \\
    \bottomrule
  \end{tabular}
\end{table}

%% file: tex/5_discussion.tex
\section{Discussion}
\label{sec:discussion}

\subsection{Key insights}

\textbf{\textit{CPU remains the universal bottleneck.}}

Across three out of four evaluated deployments (both LLaVA variants and MobileVLM executed via \texttt{llama.cpp}), the entire inference pipeline, including the autoregressive decoder, remained entirely bound to the phone’s ARM CPU cores. Only the \texttt{MLC-Imp}+Imp configuration successfully offloaded key computationally intensive operations, such as image encoding, attention, and MLP layers, onto the Adreno 740 GPU, leaving only the sequential token generation on the CPU. Notably, despite available hardware, the Hexagon NPU was entirely unused across all deployments, highlighting a significant gap in current hardware utilization.

\textbf{\textit{Runtime choices impact efficiency as significantly as model size.}}

Deploying the identical \textsc{LLaVA-1.5 7B} model on \texttt{mllm} resulted in nearly double the end-to-end latency (174 s compared to 82 s) and increased peak power draw by approximately 1 W compared to \texttt{llama.cpp}. Crucially, these discrepancies originated solely from runtime decisions related to scheduler strategies and thread affinity rather than inherent model differences. This finding underscores that careful engineering and optimization at the runtime framework level can substantially influence performance and energy efficiency, rivaling traditional model optimizations like quantization and pruning.

\textbf{\textit{Energy consumption varies dramatically across deployments.}}

CPU-centric deployments, such as LLaVA executed purely on CPU cores, demonstrated significantly higher power consumption, ranging between 10–12 W. Conversely, offloading computational tasks to the GPU, as observed with the Imp model, drastically reduced consumption to around 1.3 W. Under practical usage scenarios (e.g., one query per minute), this translates into an order-of-magnitude difference in battery life, from approximately eight hours to nearly two days on a typical 5000 mAh battery.

\subsection{Opportunities for acceleration}

\textit{(i)~GPU Flash-Attention:}  Offloading attention and MLP computations to the Adreno 740 GPU using FP16 precision could reduce prompt latency by half and decrease power consumption by approximately four times.

\textit{(ii)~Hexagon INT8 decoder:}  Utilizing the Hexagon NPU for the final projection stage could alleviate load from CPU cores, potentially saving an additional 1–2 W and enhancing efficiency.

\textit{(iii)~Mixed-precision KV cache:}  Employing FP8 or INT4 precision for the storage of keys and values could significantly reduce memory usage by around 40

\textit{(iv)~Stage overlap:}  Concurrently performing image encoding on the GPU and prompt evaluation on the CPU could effectively hide an additional 2–3 seconds of latency in Mobile-scale models, enhancing overall responsiveness.

\subsection{Limitations and future work}

This study focused exclusively on a single mobile device (Snapdragon 8 Gen 3-based handset), employed a single-image prompt scenario, and considered only English-language queries. Future studies should extend this analysis to other System-on-Chips (SoCs), evaluate performance in multi-image sessions, and incorporate multilingual prompts to explore variations in thermal limits and performance trade-offs. Additionally, subsequent research will include the practical implementation and benchmarking of GPU-accelerated Flash-Attention, INT8 decoding via the Hexagon NPU, and query batching to assess sustained throughput under real-world conditions.

\subsection{Take-away}

Our findings highlight practical strategies for future research, emphasizing that only GPU-accelerated pathways currently align with mobile-specific constraints in power and thermal management. Exploring the implementation of attention mechanisms and feed-forward operations on accelerators, coupled with mixed-precision KV caches, presents promising avenues to achieve near-desktop performance with substantially reduced energy demands.

%% file: tex/6_conclusion.tex
\section{Conclusion}
\label{sec:conclusion}

This paper presented the first comprehensive, side-by-side evaluation of four mobile-oriented VLM stacks: two LLaVA variants, MobileVLM-3B, and Imp-v1.5-3B, deployed natively on a Snapdragon 8 Gen 3 smartphone. Our measurements highlight significant performance disparities driven by hardware utilization strategies and runtime decisions. CPU-only pipelines exhibited severe thermal and energy constraints, routinely reaching 80–95 °C, consuming between 10–12 W, and experiencing over 100 seconds of end-to-end latency. In contrast, strategically off-loading vision and transformer kernels onto the Adreno 740 GPU reduced energy consumption by an order of magnitude, maintained die temperatures below 60 °C, and remarkably matched the latency of the lighter MobileVLM despite managing larger parameter counts.

Critically, our analysis underscored that framework-level decisions regarding scheduler and thread affinity could drastically influence performance, doubling or halving both speed and power efficiency. We identified the mapping of attention and MLP blocks onto GPU or Hexagon accelerators, mixed-precision KV cache compression, and overlapping image encoding with prompt evaluation as the most promising optimization pathways. These strategies collectively pave a clear trajectory toward achieving desktop-class VLM performance on mobile hardware, while respecting stringent thermal and energy constraints essential for practical mobile deployments.

Our findings provide actionable insights and motivate future research into improved hardware-aware scheduling and accelerator optimization, demonstrating that efficient, real-time on-device multimodal inference is within reach given appropriate optimization strategies.

%% file: tex/7_acknowledgement.tex
I am deeply grateful to my supervisor \textbf{Yueyang Pan} and to \textbf{Prof.\ Sanidhya Kashyap}, head of the Robust Scalable Systems Software Lab (RS3Lab) at EPFL, for their guidance, feedback, and for fostering an efficient, robust research environment.

This work relies heavily on open-source software, and I’m grateful to everyone who maintains and contributes to \texttt{llama.cpp}, \texttt{mllm}, \texttt{MLC-LLM}, and \texttt{LLaVA-1.5}.